%% file: main.tex
\documentclass{article} 
\usepackage{iclr2023_conference,times}

\input{preamble.tex}

\input{math_commands.tex}

\usepackage{hyperref}
\usepackage{url}

\title{Low-Entropy Latent Variables Hurt \\ Out-of-Distribution Performance}

\author{Nandi Schoots \& Dylan Cope \thanks{Equal contribution.} \\
King's College London and Imperial College London
}

\iclrfinalcopy 
\begin{document}

\maketitle

\input{content.tex}

\newpage

\section*{Acknowledgements}

Work done by both authors is thanks to the UKRI Centre for Doctoral Training in Safe and Trusted AI (EPSRC Project EP/S023356/1).

\bibliography{references}
\bibliographystyle{iclr2023_conference}

\newpage
\appendix
\input{appendix.tex}

\end{document}

%% file: preamble.tex
\usepackage[utf8]{inputenc} 
\usepackage[T1]{fontenc}    
\usepackage{hyperref}       
\usepackage{url}            
\usepackage{booktabs}       
\usepackage{amsfonts}       
\usepackage{nicefrac}       
\usepackage{microtype}      
\usepackage{xcolor}         

\usepackage{dsfont}
\usepackage{amsmath,amssymb,amsthm,mathtools}

\usepackage{wrapfig}
\usepackage{graphicx}
\usepackage{float}
\usepackage{enumitem}
\usepackage{subcaption}

\usepackage[section]{placeins}

%% file: math_commands.tex
\usepackage{amsmath,amsfonts,bm}

\def\eqref#1{equation~\ref{#1}}

\def\1{\bm{1}}

\DeclareMathAlphabet{\mathsfit}{\encodingdefault}{\sfdefault}{m}{sl}
\SetMathAlphabet{\mathsfit}{bold}{\encodingdefault}{\sfdefault}{bx}{n}



%% file: content.tex
\begin{abstract}
We study the relationship between the entropy of intermediate representations and a model's robustness to distributional shift. We train models consisting of two feed-forward networks end-to-end separated by a discrete $n$-bit channel on an unsupervised contrastive learning task. 
Different \textit{masking strategies} are applied after training that remove a proportion of low-entropy bits, high-entropy bits, or randomly selected bits, and the effects on performance are compared to the baseline accuracy with no mask. 
We hypothesize that the entropy of a bit serves as a guide to its usefulness out-of-distribution (OOD).  
Through experiment on three OOD datasets we demonstrate that the removal of low-entropy bits can notably benefit OOD performance. 
Conversely, we find that top-entropy masking disproportionately harms performance both in-distribution (InD) and OOD.

\end{abstract}

\section{Introduction} \label{sec:intro}

The key challenge that we seek to address is that of identifying learned features in a model's intermediate representations that are more or less likely to be robust to distributional shift. 
Our approach starts from the intuition that for high-entropy features in a model's training distribution, it will have learned a better understanding for when the feature is relevant. More precisely, it will be better at distinguishing the presence or absence of the feature across different situations. Consider a hypothetical data set containing photographs from two safari trips, where each trip contains the same people on the same safari, but driving around in different trucks. Suppose that it is useful for the given task to identify which of the two trips a given image corresponds to;  we might expect the model to be particularly good at distinguishing between the trucks. On the other hand, if a rare tree appears in exactly one photograph, the model may have learned to recognise the specific pattern of pixels in that photograph corresponding to the tree, but it might not have the capability to recognise the tree in new situations.

As models have increased in performance within the bounds of the i.i.d. assumption, recent years have seen growing interest in the OOD behaviour of machine learning systems. While many approaches have studied OOD detection or the effects of external changes to a model's training regime on OOD behaviour (e.g. domain randomization or auxiliary loss functions), to the best of our knowledge our proposal of the entropy of an intermediate representation as a guide to its effects OOD is a novel approach. In this paper we demonstrate that the removal of low-entropy representations via the masking of learned discrete bits can notably improve OOD performance. 


\section{Task and Model Description} \label{sec:setup}

\begin{figure}[t]
    \centering
    \includegraphics[width=0.95\linewidth]{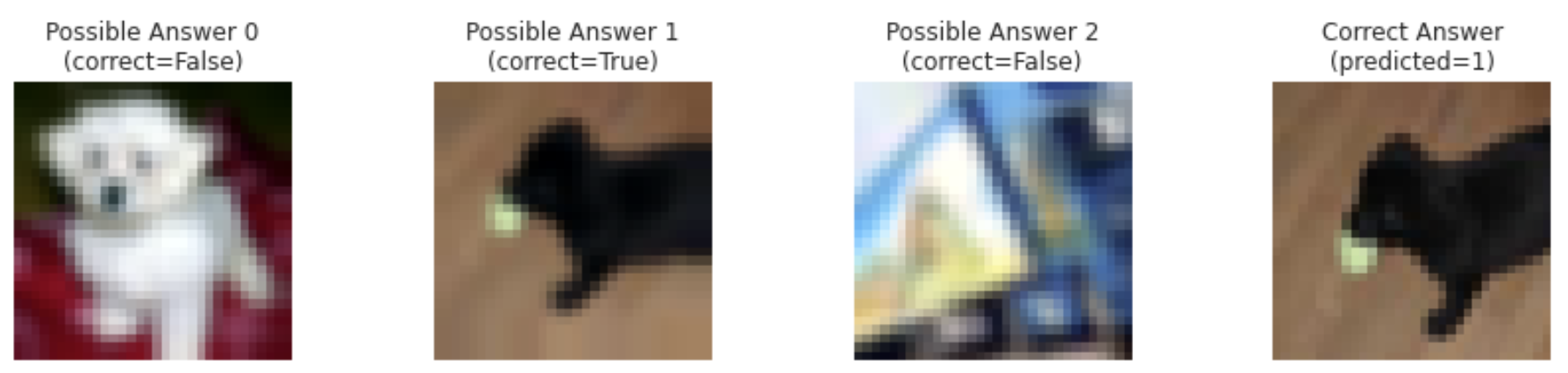}
    \caption{An example of a contrastive task ($k = 3$). For a given dataset, the distinguisher is shown  $k$ images, among which $k$-1 distractor images, and has to predict the correct image.} \label{fig:contrastive_task_example}
\end{figure}
\vspace{-0.3cm}


To learn representations of a domain we train an \textit{encoder} network to produce a representation $r$ of a given input $x^*$. This representation is given to a \textit{distinguisher} network that is tasked with identifying $x^*$ from a set of $k$ images composed of $x^*$ and $k$ - 1 \textit{distractor inputs} arranged randomly. We use the CIFAR-10 dataset \citep{Krizhevsky2009LearningImages} as the training distribution. The labels from the dataset are discarded and an unsupervised \textit{$k$-contrast task} is constructed by pairing each image with $k$ - 1 distractor images, shuffling, and giving the distinguisher $k$ inputs to choose from. The same preprocessing is later used when out-of-distribution datasets are introduced.
See Figure \ref{fig:contrastive_task_example}for an example of a contrastive task and Figure \ref{fig:contrastive_task_arch} in the Supplementary Material for the full architecture. 




It is important to note that we use a `soft-discretization' technique \citep{Foerster2016LearningLearning} on the intermediate representation $r$ such that it can be learned with gradient-descent, but each dimension can be mapped to a binary digit at test time with no loss in performance. While the use of a communication channel to discretize representations poses optimization challenges, it also provides a large benefit when it comes to computing the entropy values of each bit in the representation. The computation is reduced from approximating an integral to the simple formula for the entropy of a binary variable, as outlined in Section \ref{sec:entropy_of_bits}. This allows us to run a greater number of experiments with higher precision than if we had used continuous representations.

This unsupervised contrastive learning task was chosen as it can be easily transferred to different data distributions. A task such as image classification limits the available datasets as it requires the out-of-distribution testing data to have the same (or at least overlapping) image labels. 

\section{Entropy-based Masking} \label{sec:analysis_method}

\subsection{Entropy of Representation Bits} \label{sec:entropy_of_bits}

Each representation $r$ produced by an encoder network consists of a number of bits $|r|$, referred to as the representation length. By considering each bit at index $i$ as a random variable $B_i$ we can compute the binary entropy of the bit on a given dataset $\mathcal{D}$:
\begin{align}
    H(B_i ~|~ \mathcal{D}) = -p \log_2 p - (1 - p) \log_2 (1 - p), \quad \text{where}~ p = P(B_i = 1 ~|~ \mathcal{D}).
\end{align}

Entropy close to 1 means that the bit is 0 or 1 with roughly equal probability of $p=0.5$. Very low entropy means that the bit is either almost always 0 or almost always 1. 
We notice that for smaller representation lengths and/or few distractors the distribution tends to skew towards higher entropy bits. 
In separate experiments where we further varied representation lengths, we find that for smaller $|r|$ equal to 8, 16 or 32, all bits have entropy higher than 0.8, which makes studying bits based on entropy variation uninteresting for these representation lengths. For a visualization of these entropy values see Figure \ref{fig:dist-ent-8,16,32} in the Supplementary Material. Representation lengths of 64, 128, 256 and 512 all lead to a wide range of entropy values. A theoretical analysis of the optimal bit-entropy can be found in Section \ref{theoretical-results} of the Supplementary Material.

\subsection{Bit Masking Strategies}

In this paper we are interested in the effects of strategically `removing' parts of the model's intermediate representation, i.e. obscuring bits in $r$. It is important to note that we are only applying masking at \textit{test time}. The masking is not used to train any of the models. The mask is defined by a set \textit{masking variables} $m_i \in \{0, 1\}$ for each bit $r_i$ in the representation. The masked bit $\hat{r}_i$ is computed:
\begin{align}
    \hat{r}_i = m_i r_i + (1 - m_i) \frac{1}{2}.
\end{align}
In other words, when the masking variable $m_i = 0$ then $\hat{r}_i = 0.5$, and otherwise $\hat{r}_i = r_i$. In this paper we use three \textit{masking strategies}; Random Masking, Top-Entropy Masking, and Bottom-Entropy Masking. In order to construct a mask with any of these strategies, we define a \textit{masking proportion} $p_{\text{mask}}$ that represents the percentage of bits in $r$ that should be masked.


To construct any mask $M = \{m_1, \ldots, m_{|r|}\}$ we will need to choose $l_{\text{mask}} = \lfloor \hspace{0.1cm} p_{\text{mask}} \cdot |r| \hspace{0.1cm} \rfloor$ bits to remove. For a \textit{random mask} we draw $l_{\text{mask}}$ masking variables from $M$ at random with uniform probability and without replacement, and set them to 0, we set the remaining $|r| - l_{\text{mask}}$ variables to 1. To construct a \textit{top-entropy mask} we compute the entropy for each bit $h_i = H(B_i ~|~ \mathcal{D})$ and sort these values in descending order. We then take the bits associated with the first $l_{\text{mask}}$ entropy values (i.e. highest entropy) and set their corresponding masking variables to zero. Likewise, for the \textit{bottom-entropy mask} we take the last $l_{\text{mask}}$ bits and remove those instead.


\section{Experimental Results} \label{sec:results}

\begin{wrapfigure}{t}{0.5\textwidth}
    \centering
    \vspace{-.5cm}
    \includegraphics[width=.9\linewidth]{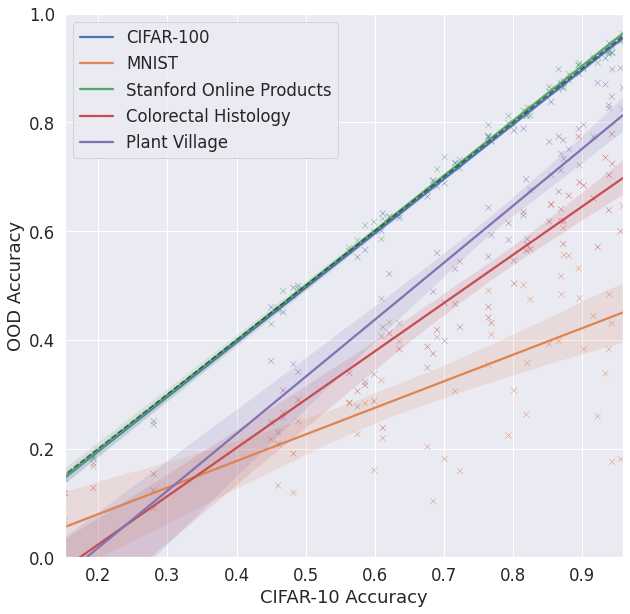}
    \caption{Accuracy of CIFAR-10 pre-trained models on OOD datasets (on the y-axis) against accuracy on CIFAR-10 (on the x-axis). The dashed line (which coincides with the green and blue lines) is the $y=x$ line.}
    \label{fig:ood_acc_shift}
    \vspace{-.5cm}
\end{wrapfigure}

We trained 54 encoder-distinguisher pairs\footnote{A sweep of 3 runs for each pair of $(|r|, k)$ plus 6 initial separate runs.} on CIFAR-10 and removed models that did not converge, resulting in 51 trained models. Models were trained with varying combinations of representation lengths and number of distractors: $(|r|, k) \in \{64, 128, 256, 512\} \times \{3, 5, 10, 20\}$. See Table \ref{tab:cifar10_acc_table} for the test accuracy statistics for the models on the $k$-contrast CIFAR-10 training distributions. See Section \ref{app:training-meth} in the Supplementary Material for a full description of the training methodology.

\begin{table}[]
    \centering
    \begin{tabular}{c|cccc}
    \toprule
                 & \multicolumn{4}{c}{Representation Length} \\[5pt]
    Training $k$ &              64    &              128   &              256   &              512   \\
    \midrule
    3            &  $0.909 \pm 0.029$ &  $0.869 \pm 0.015$ &  $0.870 \pm 0.052$ &  $0.887 \pm 0.015$ \\
    5            &  $0.797 \pm 0.026$ &  $0.688 \pm 0.077$ &  $0.759 \pm 0.131$ &  $0.820 \pm 0.166$ \\
    10           &  $0.866 \pm 0.103$ &  $0.579 \pm 0.018$ &  $0.643 \pm 0.231$ &  $0.736 \pm 0.171$ \\
    20           &  $0.662 \pm 0.170$ &  $0.538 \pm 0.230$ &  $0.532 \pm 0.380$ &  $0.481 \pm 0.337$ \\
    \bottomrule
    \end{tabular}
    \vspace{0.3cm}
    \caption{Accuracy on CIFAR-10 test set of trained models with different $k$ and $|r|$ values.}
    \label{tab:cifar10_acc_table}
    \vspace{-0.4cm}
\end{table}

To evaluate the effects of distributional shifts we test our 51 trained models on the CIFAR-100 \cite{Krizhevsky2009LearningImages}, Stanford Online Products \cite{Song2016DeepEmbedding}, Colorectal Histology \cite{Kather2016Multi-classHistology}, Plant Village \cite{Hughes2015AnCrowdsourcing}, and MNIST \cite{LeCun1999MNISTDatabase} datasets. 


In Figure \ref{fig:ood_acc_shift} we demonstrate the shift in performance that results from applying the models to the new datasets. Following Taori et al. \citep{Taori2020MeasuringClassification}, plotting the relationship between InD and OOD performance in this manner allows us to study distributional shift while controlling for the variations in initial accuracy. The $y=x$ line is plotted with a black dashed line, however, it is obscured by the regression lines for CIFAR-100 and Stanford Online Products. This tells us that there is no distributional shift for these datasets, i.e. no loss in performance.  For this reason, we drop these datasets from all further out-of-distribution analysis. For the other datasets, we see in order of increased degradation: Plant Village, Colorectal Histology, and MNIST.

\subsection{Analysis of Masking Effects In-Distribution} \label{sec:masking_ind}

Before moving onto the out-of-distribution case, we first examine the effects of applying the different masking strategies to the models that we trained on CIFAR-10, with the CIFAR-10 test data. For each of the 51 successfully trained models we evaluated the accuracy without any masking, and with each of the different masking strategies for masking proportions between 0.15 and 0.5 at 0.05 intervals. We found that for any masking proportion, removing the top-entropy bits is more damaging to accuracy than masking out bottom-entropy bits. In light of general insights from information theory, this result is not too surprising. The highest entropy bits necessarily convey the most information, and so it follows that their removal should lead to the largest drop in performance. 

In general, we did not expect any of the masking strategies to provide a benefit when applied within the training distribution. Yet, we saw that with a small masking proportion (around $p_{\text{mask}} < 0.3$) we see an \textit{increase} in accuracy for low-entropy and random masks. 
Our initial hypothesis was that the masking may be `undoing' overfitting to the training set. But for each of the trained models we have verified that there is no overfitting (see Section \ref{app:overfitting-analysis} in the Supplementary Material for a visualization).




\subsection{Analysis of Masking Effects Out-of-Distribution (OOD)} \label{sec:masking_ood}


In order to understand the effects of masking on accuracy in the OOD setting we measure the \textit{mean change in accuracy} of a masking strategy under various circumstances. We also report the standard deviations associated with these estimates. As in the case of in-distribution masking we evaluated the masking strategies for a sweep of masking proportions between 0.15 and 0.5 at 0.05 intervals. We cut-off the maximum masking proportion $p_{\text{mask}} \leq 0.25$ for all further analysis as beyond that threshold masking has an almost universally negative effect. The overall mean accuracy changes can be seen in Table \ref{tab:exp_acc_change_mask_overall}. We see that masking the bottom-entropy or random bits produces the highest increase, albeit with a large variance.

\begin{figure*}[t]
    \centering
    \includegraphics[width=1\textwidth]{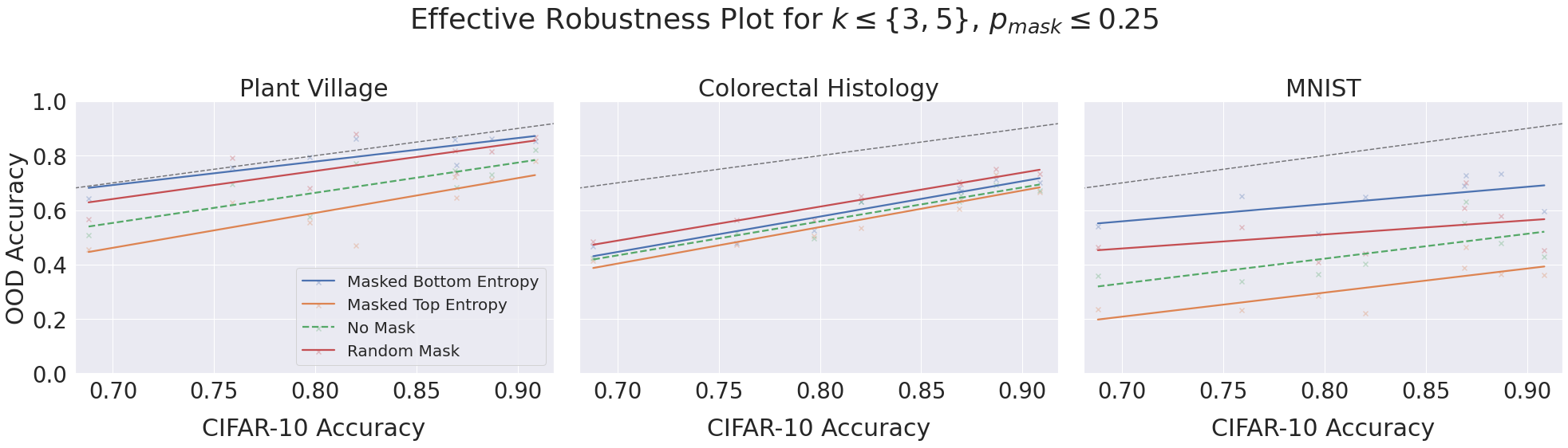}
    \caption{Effective robustness plots for low-$k$ models. $y=x$ shown as black dashed line.}
    \label{fig:eff_robustness_k35}
\end{figure*}

This variance can be understood and disentangled by separating the low-$k$ models from the high-$k$ models. What we see is that the benefits of bottom-entropy masking are more prevalent for low-$k$ models. This is visualized in Figure \ref{fig:eff_robustness_k35} where we illustrate the \textit{effective robustness} of each of the masking strategies on the three OOD datasets. 
In the Supplementary Material Section \ref{app:values-k-r} we include plots for all values of $k$ and $p_{\text{mask}}$ that we tested. 
Effective robustness is a concept introduced by \cite{Taori2020MeasuringClassification} as a way to understand the efficacy of a method for increasing robustness to distributional shift. By plotting the baseline regression line for unaltered models with differing in-distribution accuracy values on the diagram we can observe whether a proposed robustness method moves towards the $y=x$ line (i.e. no degradation). Crucially, with these plots, we are able to account for each model's performance on the training distribution. Hence, despite the large variance in the performance of models trained across various $k$ and $|r|$ values\footnote{Accuracy ranging between 0.65 and 0.95 for even the high-performing low-$k$ models, as shown in the $x$-axes of Figure \ref{fig:eff_robustness_k35}.}, we are able to discern the effects of the masking interventions.

In our case, we see that -- as is consistent with previous results -- for each dataset the top-entropy masking moves below the dashed green line showing the baseline unmasked models. On the other hand, the random masking and bottom-entropy masking lines move closer to $y=x$ (as compared to the no masking lines). For Plant Village we see that almost all of the in-distribution accuracy is recovered. For MNIST we find the most substantial jump, and the largest benefit of bottom-entropy over random masking.

\begin{table*}[t]
\centering
\begin{tabular}{lcccc}
\toprule
&       CIFAR-10 & Colorectal Histology &          MNIST &   Plant Village \\
\midrule
Masked Bottom Entropy &    $1.6 \pm 8.0$ &      $-2.0 \pm 14.3$ &   $9.4 \pm 15.6$ &    $3.0 \pm 23.7$ \\
Masked Top Entropy    &  $-4.3 \pm 21.4$ &      $-7.8 \pm 19.0$ &  $-16.6 \pm 5.3$ &  $-18.5 \pm 21.8$ \\
Random Mask           &   $2.5 \pm 12.3$ &       $3.4 \pm 10.9$ &   $4.2 \pm 13.7$ &    $2.1 \pm 19.6$ \\
\bottomrule
\end{tabular}
\caption{Mean accuracy shift (in percentage points) after masking with each strategy. After running paired t-tests we find that all of these accuracy shifts are statistically significant (with $p=0.05$).} \label{tab:exp_acc_change_mask_overall}
\end{table*}



\section{Related Work} \label{sec:related_work}

Our work adds to the toolkit of methods to aid in understanding and improving robustness to distributional shift, which for example includes forms of data augmentation \cite{DBLP:conf/iccv/HendrycksBMKWDD21} and abstaining from making a prediction in the face of uncertainty \cite{DBLP:conf/icmla/ThulasidasanTDC21}. For a general overview of  problems and methods in OOD robustness see \cite{Shen2015TowardsSurvey}.

Below we reference some notable entropy-based methods that have a \textit{different purpose} than improving OOD robustness.
\cite{https://doi.org/10.48550/arxiv.2002.10657} use low entropy (or ``rare'') signals to analyze the extent to which a model is overfitted to the training distribution. 
Entropy-based methods have also been used widely in the adjacent problem of OOD detection. For example, predictive entropy measures the uncertainty of the prediction of a sample given a training distribution and is used to calculate the extent to which a sample is OOD \cite{ood-detection-entropy}.
However, we apply entropy in an entirely different context, namely, we calculate the entropy of \emph{latent variables} to estimate how robust they will be to distributional shift.
Relative entropy (KL-divergence) is a popular measure and is notably used in the Bits-Back method \cite{Hinton-van-Camp}, \cite{DBLP:journals/corr/abs-2010-01185} 
to calculate the optimal compression rate in latent variables. 
Images that are traditionally compressed by a variational auto-encoder have now been compressed with code-length close to this theoretical optimum \cite{DBLP:journals/corr/abs-2010-01185}.

Contrastive representation learning takes many forms; in computer vision alone there are many approaches for applying deep learning to multiple inputs and producing representations to distinguish between them; see \cite{Jaiswal2020ALearning} for a review.  To our knowledge, there are no existing suitable state-of-the-art (SOTA) methods for OOD robustness in contrastive learning to benchmark our proposals against. 


\section{Conclusion} \label{sec:conc}




In this paper we have investigated the out-of-distribution effects of using different post-hoc strategies to remove bits from discrete intermediate representations in an unsupervised contrastive learning task. We have studied how the difficulty of the task (more distractors) impacts the entropy distribution of the learned representations and shown that removing low-entropy bits can improve the performance of models out-of-distribution (Section \ref{sec:masking_ood}), notably almost entirely restoring in-distribution performance for one of our datasets (see Figure \ref{fig:eff_robustness_k35}). 
However, the results also present mysteries that prompt further experiments and analysis. At the time of writing, we do not have a clear understanding of why the removal of bits within the training distribution should increase performance, as we would expect the encoder to learn an optimal protocol. 

Next, there is a need for a deeper understanding of the conditions in which our results hold. Within our experimentation, we found that the effect (of harm from low-entropy features OOD) was less pronounced for models trained on the more difficult tasks (higher numbers of distractors). From our data, it is unclear if this relationship represents something fundamental or if it is a side-effect of these models generally performing to a lower standard. One of the most important avenues of further work is in testing if other systems built on top of the learned representations in this paper inherit the same OOD robustness under low-entropy masking.

%% file: appendix.tex
\section{Network Architectures}\label{app:network-arch}
\begin{wrapfigure}{r}{0.55\textwidth}
    \centering
    \includegraphics[width=0.95\linewidth]{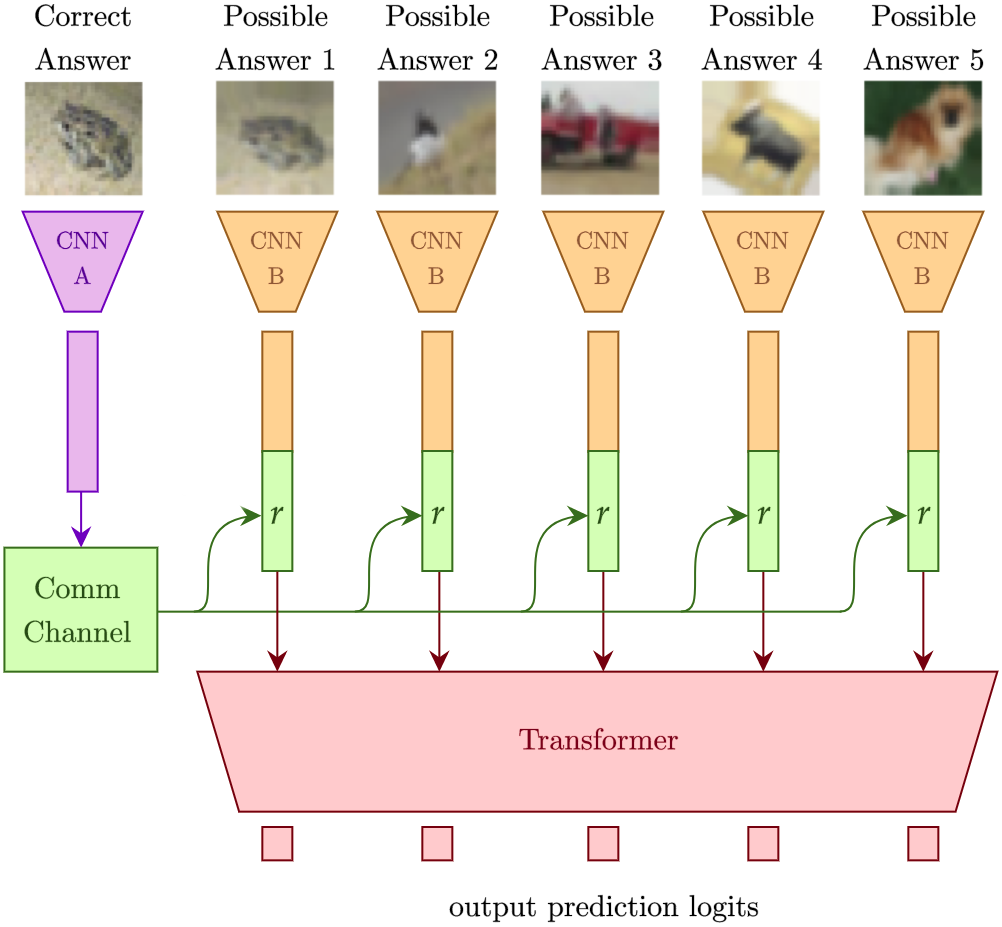}
    \caption{Architecture diagram ($k=5$). The encoder is shown as the purple and green components, and the distinguisher is the orange and red.}
    \label{fig:contrastive_task_arch}
    \vspace{-0.4cm}
\end{wrapfigure}

The encoder network is composed of a convolutional network (CNN) that takes a $32\times 32 \times 3$ dimensional tensor as input (CNN$_A$ in Figure \ref{fig:contrastive_task_arch}), followed by: a $3 \times 3$ convolutional layer with 64 filters and ReLU activation; two $3 \times 3$ convolutional layers with 64 filters, ReLU activation, and a stride-length of 2; a flatten layer; and finally a dense layer without any activation that projects into $\mathds{R}^{|r|}$, where $|r|$ is a hyperparameter controlling the `representation length' of $r$.
Next, between the encoder and the distinguisher, there is a \textit{discretize/regularize unit} \citep{Foerster2016LearningLearning}. Following the literature in which this component was developed, we will refer to this as a \textit{communication channel} (see in green in Figure \ref{fig:contrastive_task_arch}). The channel is a differentiable unit that, during training, `soft discretizes' activations passed through it by applying Gaussian white noise (GWN) and a sigmoid function. Then at test time we `hard discretize' the activations by passing through a sigmoid function and emitting 0 if the result is less than 0.5 and 1 otherwise. This enables the end-to-end learning of a discrete representation via backpropagation from the output of the distinguisher. We configure the channel with a fixed GWN standard deviation of 0.5 during training.

The distinguisher network is composed of another convolutional network (CNN$_B$ in Figure \ref{fig:contrastive_task_example}) with exactly the same input and layers as the CNN in the encoder (initialized separately and no parameter sharing), except projecting to a fixed embedding size of 128. This CNN is shared for each of the `possible answer' images, producing embeddings that are each concatenated with the representation $r$ from the encoder (i.e. the output of the communication channel) and fed into a transformer network \citep{Vaswani2017AttentionNeed} as tokens. The transformer is composed of two self-attention encoder layers with 3 heads of dimension 64, and a dropout rate of 0.1. After the transformer layers each token is projected onto a single dimension without activation. This is then taken as the log-probability (logit) that the corresponding possible answer is correct. The networks are trained together with a sparse categorical crossentropy loss on these logits and the index of the correct answer. The use of a transformer and a shared encoder for the input images means that a model trained, for example, on a 3-contrast dataset ($k=3$) can be tested on a 5-contrast dataset without any modification.

\section{Theoretical Result}\label{theoretical-results}

Consider the following abstracted and idealized version of the contrastive learning game. 
An encoder receives an input, and communicates features in that input via bits. A distinguisher has to identify the original input from a set of $k$ (distractor) inputs, based on the communicated features. The encoder and distinguisher win if the distinguisher correctly identifies the original input. 
The encoder and distinguisher need to decide on a communication protocol before playing the game. Each bit corresponds to one feature. The encoder sends a 1 if a given feature is present and a 0 otherwise.

The question we're answering in this section is: what is the optimal feature occurrence (or bit entropy) for a feature when the encoder can choose $b$ bits, and the distinguisher has to choose between $k$ inputs.

Below we calculate that the optimal strategy is to use $l$ independent features that are each present in exactly half of the images. The chance of the receiver picking out the right image depends on $k$.

For this calculation we will assume the encoder can choose $b=2$ bits, i.e. can communicate two features $x$ and $y$.
Let $f_x$ and $f_y$ be the frequency of respectively feature $x$ and feature $y$ in the dataset. To answer the question we will calculate the values of $f_x$ and $f_y$ that maximize the chance of winning.

Let $c_x$ be the random variable that represents: the correct input has feature $x$, and $c_y$ the variable that represents: the correct input has feature $y$. We assume that these variables are independent.
Let $v$ be the random variable that represents the number of inputs in the set of $k$ inputs that the distinguisher gets to see, that have both feature $x$ and feature $y$.

\begin{equation*}
    \begin{split}
       P(\text{win})  = \quad & P(\text{win} |c_x , c_y) \cdot P(c_x , c_y) \\
                 +&  P(\text{win} |c_x , c_{\neg y}) \cdot P(c_x , c_{\neg y}) \\
                 +& P(\text{win} |c_{\neg x} , c_y) \cdot P(c_{\neg x} , c_y) \\
                 +& P(\text{win} |c_{\neg x} , c_{\neg y}) \cdot P(c_{\neg x} , c_{\neg y}) 
    \end{split}
\end{equation*}

Note that $P(c_x, c_y) = f_x \cdot f_y$. Below we calculate that \[ P(\text{win}|c_x, c_y) = \Sigma_{v=1}^k \frac1v \cdot (f_x f_y)^{v-1} \cdot (1 - f_x f_y)^{k-v} \cdot \binom{k-1}{v-1}. \]

To do so we introduce one more helper variable $\tilde{v}$ which represents the number of inputs in the set of $k$ inputs that the distinguisher gets to see, that have both feature $x$ and feature $y$, but excluding the correct input.

We now calculate
\begin{equation*}
    \begin{split}
 P(\text{win}|c_x, c_y) 
& = \Sigma_{v=1}^k P(\text{win} |c_x , c_y, V=v) \cdot P(\tilde{V} = v-1| c_x , c_y) \\
& = \Sigma_{v=1}^k P(\text{win} |c_x , c_y, V=v) \cdot P(\tilde{V} = v-1) 
 \end{split}
\end{equation*}

Note that $P(\text{win} |c_x , c_y, V=v) = \frac1v$ and 
\[P(\tilde{V} = v-1) = (f_x f_y)^{v-1} \cdot (1 - f_x f_y)^{k-1-(v-1)} \cdot \binom{k-1}{v-1} \]

Hence 
\[ P(\text{win}|c_x, c_y) 
 = \Sigma_{v=1}^k \frac1v \cdot (f_x f_y)^{v-1} \cdot (1 - f_x f_y)^{k-v} \cdot \binom{k-1}{v-1}. \]

Applying the Bionomial theorem gives us the following equality 
\begin{equation*}
    \begin{split}
 P(\text{win}|c_x, c_y) P(c_x, c_y) 
&= f_x f_y \cdot \Sigma_{v=1}^k \frac1v \cdot (f_x f_y)^{v-1} \cdot (1 - f_x f_y)^{k-v} \cdot \binom{k-1}{v-1} \\
&= \Sigma_{v=1}^k \frac1v \cdot (f_x f_y)^{v} \cdot (1 - f_x f_y)^{k-v} \cdot \binom{k-1}{v-1} \\
&= \frac1k \Sigma_{v=1}^k  (f_x f_y)^{v} \cdot (1 - f_x f_y)^{k-v} \cdot \binom{k}{v} \\
&= \frac1k \left(  (f_x f_y + 1 - f_x f_y)^k - (1 - f_x f_y)^k   \right)\\
&= \frac1k \left(  1 - (1 - f_x f_y)^k   \right) \\
&= \frac1k - \frac1k (1 - f_x f_y)^k 
 \end{split}
 \end{equation*}

We can write similar equations for $c_{\neg x}$ and $c_{\neg y}$ and combining them results in 
\begin{equation*}
    \begin{split}
    P(\text{win}) = \frac{4}k - \frac1k \big( & (1- f_x \cdot f_y)^k \\
    + & (1- f_x \cdot (1-f_y))^k \\
    + & (1- (1-f_x) \cdot f_y)^k \\
    + & (1- (1-f_x) \cdot (1-f_y))^k \big).
 \end{split}
\end{equation*}

More generally, for arbitrary number of bits $b$ and feature frequencies $f_1, \ldots, f_b$  we find
\[P(\text{win}) = \frac{2^b}k - \frac1k \left( (1- f_1 \cdots f_b)^k 
    + (1- (1-f_1) f_2 \cdots f_b)^k 
    +  \ldots 
    + (1- (1-f_1)\cdots (1-f_b))^k \right)\]

The derivative of $P(\text{win})$ with respect to $f_1$ is
\begin{equation*}
    \begin{split}
    \frac{\partial P(\text{win})}{\partial f_1} = \quad & f_2 \cdots f_b(1- f_1 \cdots f_b)^{k-1}  \\
    - & f_2 \cdots f_b (1- (1-f_1) f_2 \cdots f_b)^{k-1} \\
    + & \ldots \\
    - & (1-f_2)\cdots (1-f_b) (1- (1-f_1)\cdots (1-f_b))^{k-1} 
 \end{split}
\end{equation*}
When $f_1=0.5$ the components with a factor of $f_1$ compensate for the ones with a factor of $(1-f_1)$, and so the derivative is 0 for $f_1 = 0.5$. Deriving with respect to other feature values gives analogues results.
That is, one optimal feature occurrence value for maximizing $P(\text{win})$ is $0.5$.

\section{Training Methodology}\label{app:training-meth}

In order to prevent overfitting and the representation of `trivial features' (e.g. specific pixel values) in the representations, during training we use a stack of image augmentation layers independently applied prior to each image encoder. This involves a random rotation of up to 0.1 radians, a random contrast shift of up to 10\%, a random translation of up to 10\% along both axes, and a random zoom of up to 10\% (all with a nearest-neighbour filling of blank pixels).

The models were optimized using Adam \citep{Kingma2015Adam:Optimization} with a learning rate of 0.001. The batch size used for training was dependent on the number of distractors, and each epoch iterated through the entire training dataset. See Table \ref{tab:cifar10_acc_table} for the full breakdown of test accuracy values for trained models, i.e. the mean and standard deviations for the proportion of occasions where the distinguisher was correctly able to identifier $x^*$ by using $r$. 

All of the code was implemented with Tensorflow 2 \citep{Abadi2015TensorFlow:Systems} and datasets were pulled from Tensorflow Datasets\footnote{The license for these datasets can be found at: \url{https://github.com/tensorflow/datasets/blob/master/LICENSE}} (TFDS) \citep{TFDevs2022TensorFlowDatasets}. CIFAR-10 was split into the default TFDS training and test sets (50,000 training images and 10,000 test images). Training and analysis were performed with an NVIDIA RTX 3090 GPU.

We trained 54 independent encoder-distinguisher pairs\footnote{A sweep of 3 runs for each pair of $(|r|, k)$ plus 6 initial separate runs.} for 10 epochs on CIFAR-10 and removed models that did not converge (as defined by not reaching an 80\% drop in loss), resulting in 51 trained models (taken as the best performing checkpoint). Models were trained with varying combinations of representation lengths and number of distractors: $(|r|, k) \in \{64, 128, 256, 512\} \times \{3, 5, 10, 20\}$. We also trained models with representation lengths 8, 16 and 32, visualizations of which can be found in Figure \ref{fig:dist-ent-8,16,32}, which we discarded because their bit entropies were too homogeneous to meaningfully study the effect of masking out low versus high entropy bits.
\begin{figure}[H]
    \centering
    \includegraphics[width=.6\textwidth]{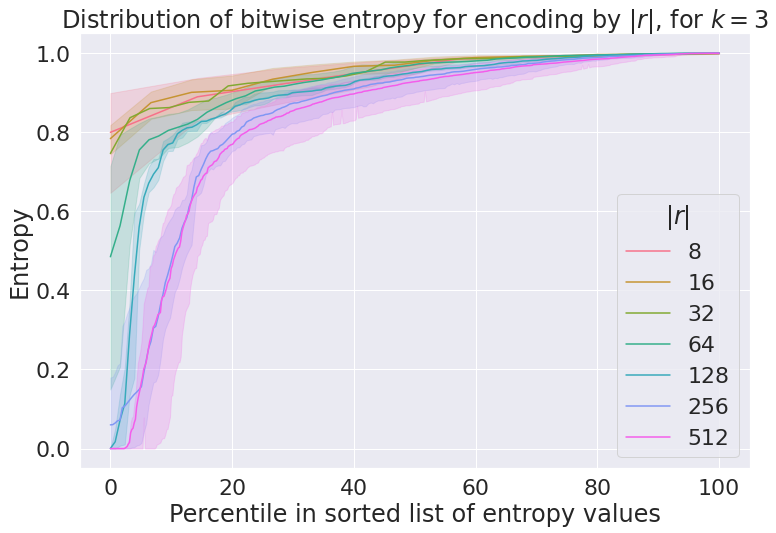}
    \caption{The x-axis represents entropy percentile of bits in the representation. The y-axis shows the entropy values of bits (measured on CIFAR-10). In other words, we take the list of bits and sort them by entropy, and then plot the sorted line as percentiles in order to compare the distributions of different lengths. The translucent regions show the error bars from various training runs. We can see that for lower $|r|$ values, the entropy distributions do not tend to go below 0.8.}
    \label{fig:dist-ent-8,16,32}
\end{figure}

\section{Experiments}

The code for the experiments can be found at the following repository: 
\textbf{[URL removed to preserve anonymity]}







\subsection{Overfitting Analysis}\label{app:overfitting-analysis}

\begin{figure}[H]
    \centering
    \includegraphics[width=.6\textwidth]{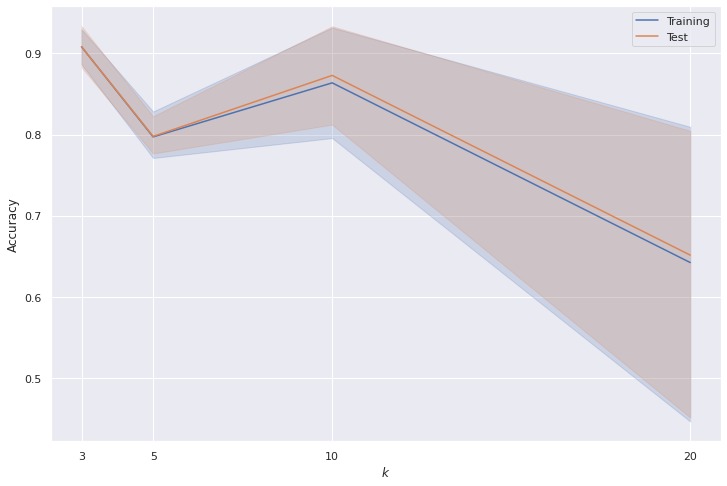}
    \caption{For different values of $k$ the blue line shows the training accuracy and the orange line shows the test accuracy.}
    \label{fig:overfitting}
\end{figure}

In Figure \ref{fig:overfitting} we see that the test and training accuracies are very similar (with the test accuracy even being slightly higher) and so no overfitting has happened.

\subsection{OOD Accuracy Change for all Masking Proportions and all values of $k$ and $|r|$}\label{app:values-k-r}

\begin{figure}[h]
    \centering
    \includegraphics[width=1\textwidth]{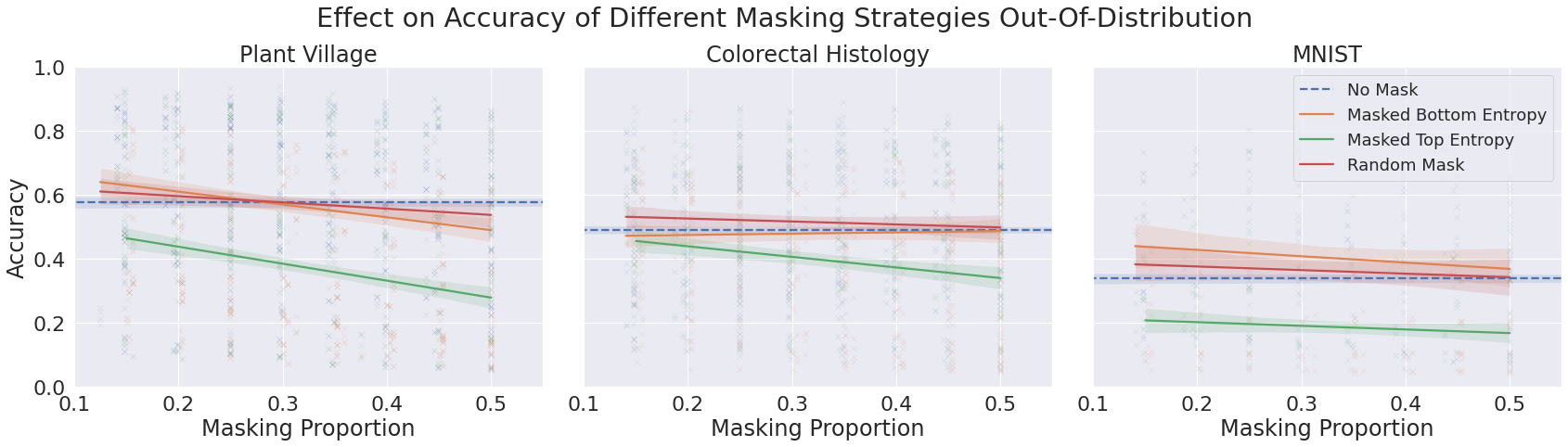}
    \caption{The y-axis represents the accuracy and the x-axis the masking proportion. Different masking strategies are represented by different colors.}
    \label{fig:OOD-accuracies-bigger-font}
\end{figure}

\begin{figure}
\centering
\begin{subfigure}[h]{1\textwidth}
    \centering
    \includegraphics[width=1\textwidth]{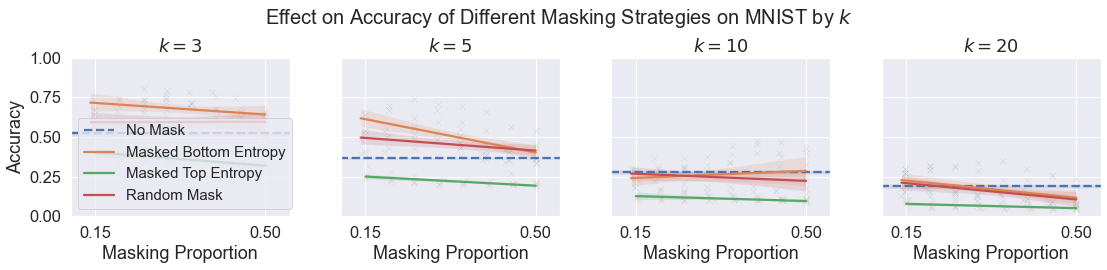}
\end{subfigure}
\begin{subfigure}[h]{1\textwidth}
    \centering
    \includegraphics[width=1\textwidth]{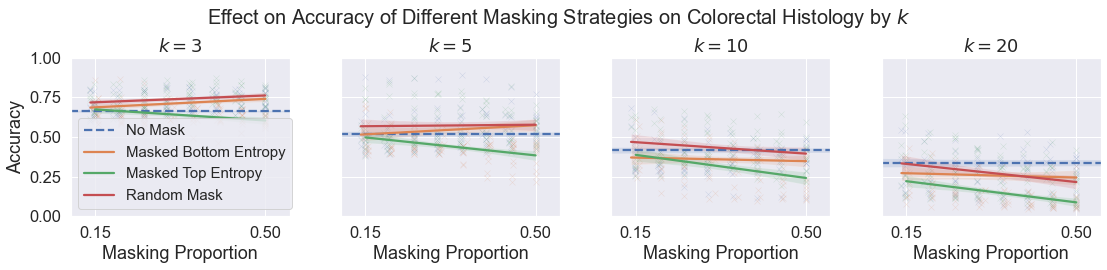}
\end{subfigure}
\begin{subfigure}[h]{1\textwidth}
    \centering
    \includegraphics[width=1\textwidth]{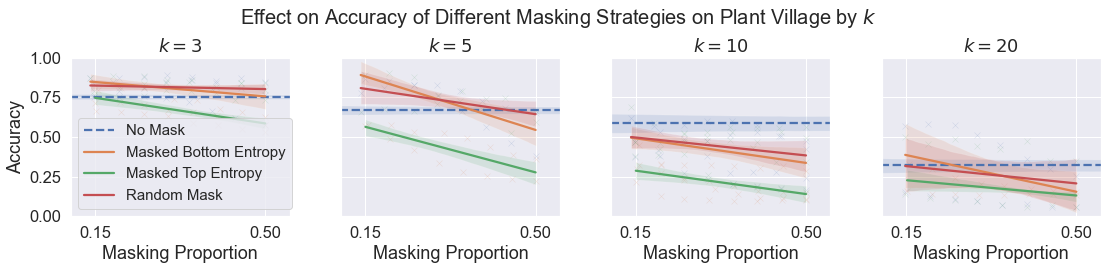}
\end{subfigure}
\caption{The y-axis shows the accuracy and the x-axis shows different masking proportions. Masking strategies are indicated by color.}
\label{fig:OOD-accuracies-by-k}
\end{figure}

\begin{figure}
\centering
\begin{subfigure}[h]{1\textwidth}
    \centering
    \includegraphics[width=1\textwidth]{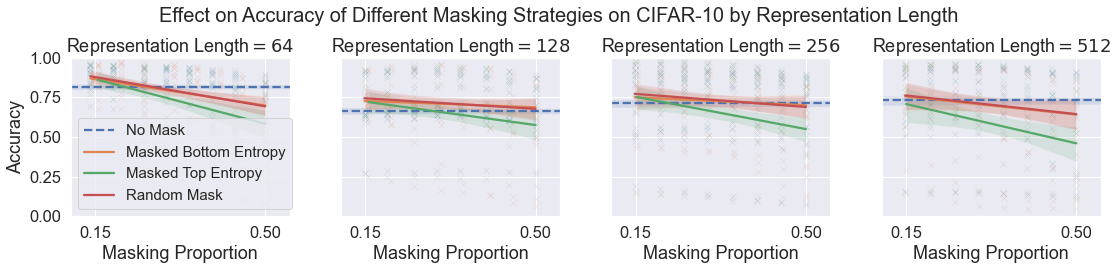}
\end{subfigure}
\begin{subfigure}[h]{1\textwidth}
    \centering
    \includegraphics[width=1\textwidth]{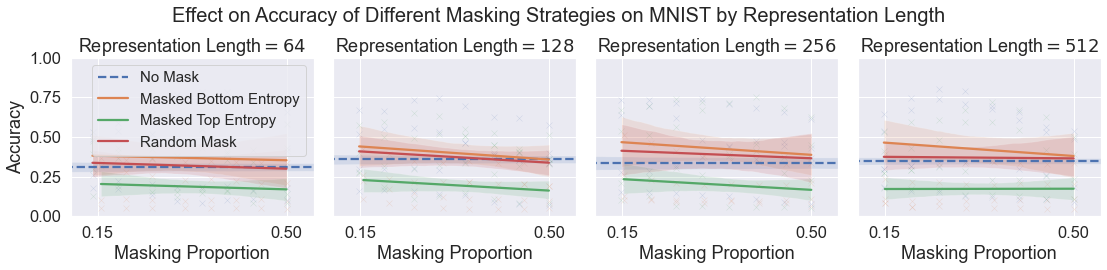}
\end{subfigure}
\begin{subfigure}[h]{1\textwidth}
    \centering
    \includegraphics[width=1\textwidth]{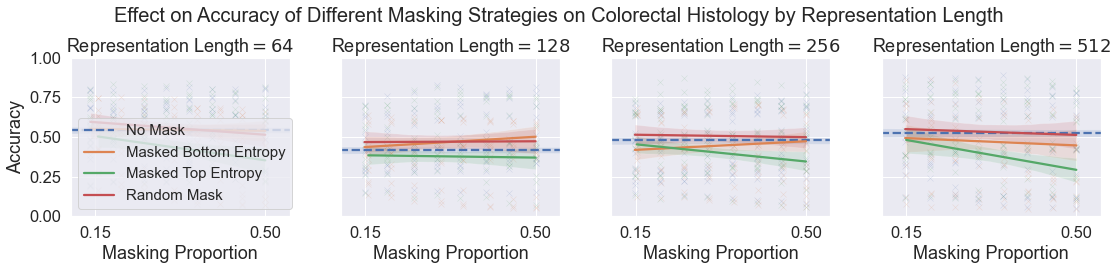}
\end{subfigure}
\begin{subfigure}[h]{1\textwidth}
    \centering
    \includegraphics[width=1\textwidth]{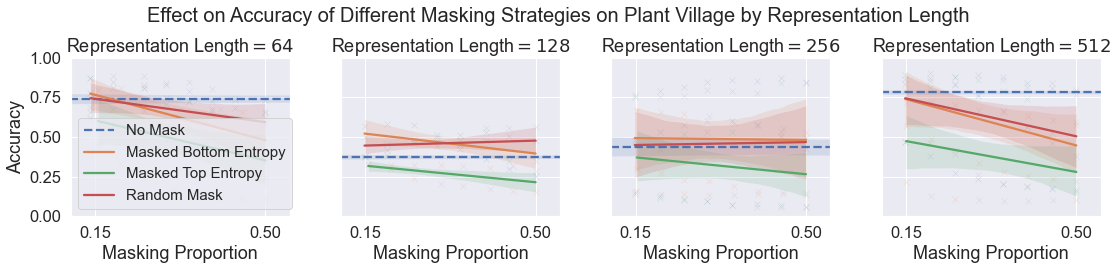}
\end{subfigure}
\caption{The y-axis shows the accuracy and the x-axis shows different masking proportions. Masking strategies are indicated by color.}
\label{fig:OOD-accuracies-by-r}
\end{figure}

Figure \ref{fig:OOD-accuracies-bigger-font} shows the OOD accuracies for each dataset (using the data of all the values of $k$ and all the analysed representation lengths). Figure \ref{fig:OOD-accuracies-by-k} shows the accuracies for each dataset and each value of $k$. Figure \ref{fig:OOD-accuracies-by-r} shows the accuracies for each dataset and each representation length $|r|$.

 \FloatBarrier

\subsection{OOD Mean Accuracy Change From Masks}

\begin{figure}[h]
    \centering
    \includegraphics[width=1\textwidth]{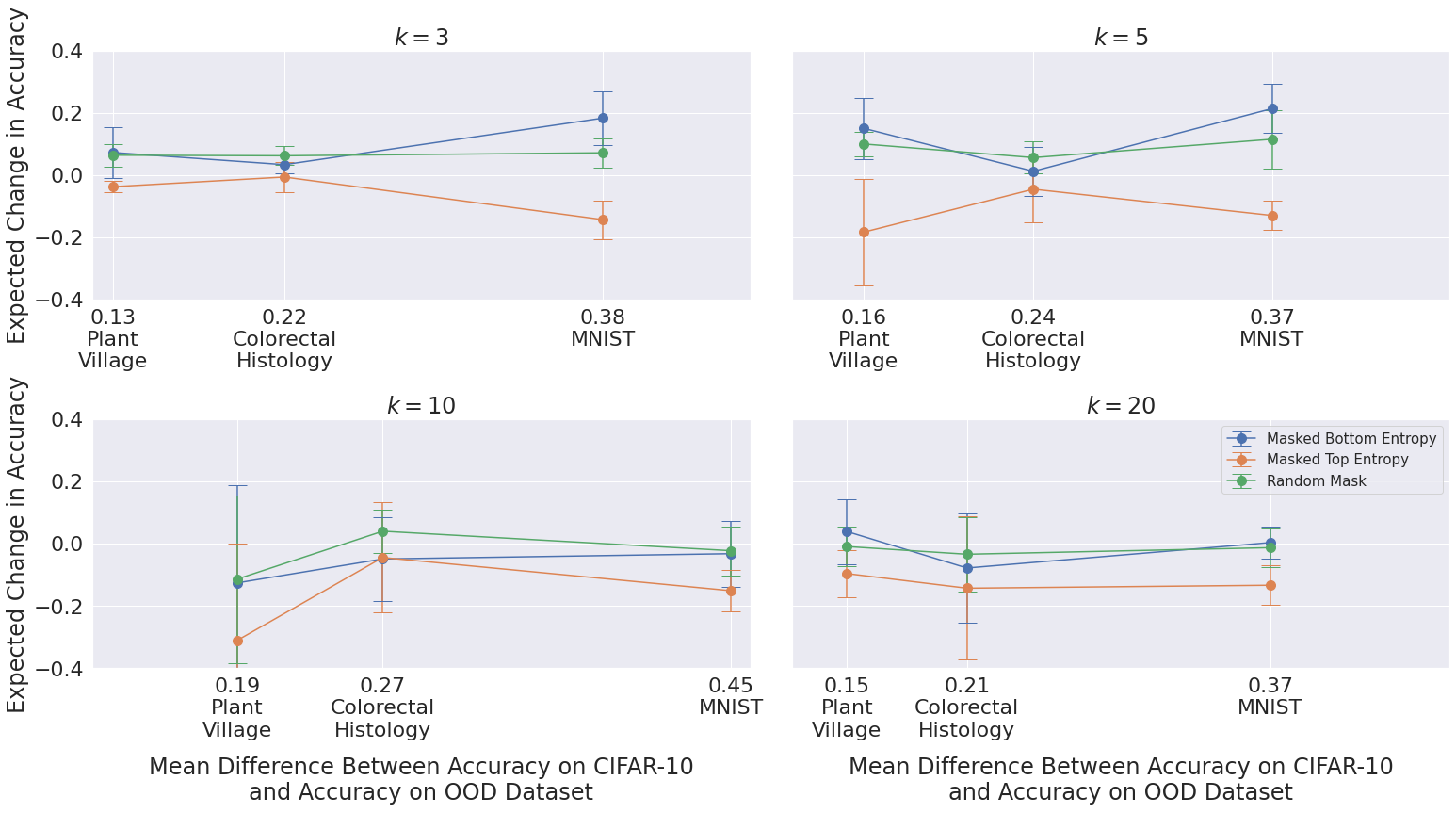}
    \caption{Mean change in accuracy when apply each masking strategy to each model  varying by $k$ on the OOD datasets. Standard deviations denoted with error bars.}
    \label{fig:exp_acc_change_plots}
\end{figure}

The tables in this section are the same as Table \ref{tab:exp_acc_change_mask_overall} in Section \ref{sec:behaviour_ood}, except separated by different values of $k$. Figure \ref{fig:exp_acc_change_plots} is a visualisation of the data along with the `distance out-of-distribution' for each $k$ value.

\begin{table}[H]
\centering
\caption{Mean change in accuracy (in percentage points) ($k=3$, $p_{mask}\approx 0.25$)}
\begin{tabular}{lllll}
\toprule
Dataset &       CIFAR-10 & Colorectal Histology &            MNIST &   Plant Village \\
Strategy              &                &                      &                  &                 \\
\midrule
Masked Bottom Entropy &  $2.8 \pm 1.4$ &        $2.8 \pm 2.3$ &   $16.6 \pm 7.5$ &   $8.1 \pm 5.5$ \\
Masked Top Entropy    &  $4.0 \pm 1.4$ &       $-0.4 \pm 4.3$ &  $-12.7 \pm 5.2$ &  $-2.9 \pm 1.5$ \\
Random Mask           &  $4.0 \pm 1.4$ &        $5.8 \pm 3.0$ &    $6.5 \pm 3.9$ &   $6.2 \pm 2.5$ \\
\bottomrule
\end{tabular}
\end{table}

\begin{table}[H]
\centering
\caption{Mean change in accuracy (in percentage points) ($k=5$, $p_{mask}\approx 0.25$)}
\begin{tabular}{lllll}
\toprule
Dataset &         CIFAR-10 & Colorectal Histology &            MNIST &    Plant Village \\
Strategy              &                  &                      &                  &                  \\
\midrule
Masked Bottom Entropy &  $4.5 \pm 3.5$ &        $0.7 \pm 7.6$ &   $22.2 \pm 8.0$ &    $13.8 \pm 9.1$ \\
Masked Top Entropy    &  $2.7 \pm 6.3$ &       $-3.4 \pm 8.5$ &  $-12.3 \pm 4.6$ &  $-13.3 \pm 15.0$ \\
Random Mask           &  $5.8 \pm 3.0$ &        $5.1 \pm 4.9$ &    $9.5 \pm 7.4$ &     $9.8 \pm 4.9$ \\
\bottomrule
\end{tabular}
\end{table}

\begin{table}[H]
\centering
\caption{Mean change in accuracy (in percentage points) ($k=10$, $p_{mask}\approx 0.25$)}
\begin{tabular}{lllll}
\toprule
Dataset &          CIFAR-10 & Colorectal Histology &            MNIST &     Plant Village \\
Strategy              &                   &                      &                  &                   \\
\midrule
Masked Bottom Entropy &    $2.2 \pm 5.8$ &      $-4.0 \pm 12.2$ &   $-3.3 \pm 9.2$ &   $-3.0 \pm 24.0$ \\
Masked Top Entropy    &  $-3.9 \pm 19.0$ &      $-4.8 \pm 17.3$ &  $-15.4 \pm 7.4$ &  $-21.3 \pm 26.5$ \\
Random Mask           &    $3.5 \pm 9.1$ &        $3.9 \pm 5.8$ &   $-1.1 \pm 5.5$ &   $-3.1 \pm 20.7$ \\
\bottomrule
\end{tabular}
\end{table}

\begin{table}[H]
\centering
\caption{Mean change in accuracy (in percentage points) ($k=20$, $p_{mask}\approx 0.25$)}
\begin{tabular}{lllll}
\toprule
Dataset &          CIFAR-10 & Colorectal Histology &            MNIST &    Plant Village \\
Strategy              &                   &                      &                  &                  \\
\midrule
Masked Bottom Entropy &   $-0.6 \pm 5.7$ &      $-7.3 \pm 17.5$ &    $0.7 \pm 4.9$ &   $-2.4 \pm 17.2$ \\
Masked Top Entropy    &  $-8.6 \pm 20.7$ &     $-13.6 \pm 22.2$ &  $-12.9 \pm 6.4$ &  $-12.0 \pm 12.4$ \\
Random Mask           &  $-0.5 \pm 12.0$ &      $-2.6 \pm 10.3$ &   $-0.6 \pm 6.1$ &    $-0.8 \pm 8.2$ \\
\bottomrule
\end{tabular}
\end{table}